\documentclass[review]{elsarticle}

\usepackage{lineno,hyperref}
\usepackage{amsmath}
\journal{Knowledge-Based Systems}
\usepackage{booktabs}
\usepackage{graphicx}
\usepackage[lofdepth,lotdepth]{subfig} 
\usepackage{multirow}

\usepackage[textwidth=12cm]{geometry}
\usepackage{blindtext}
\usepackage{color}
\begin{document}
\begin{sloppypar}
\bibliographystyle{elsarticle-num}
\begin{frontmatter}
\title{Inferring Substitutable and Complementary Products with Knowledge-Aware Path Reasoning based on Dynamic Policy Network}

\author[label1]{Zijing Yang}
\ead{51194506043@stu.ecnu.edu.cn}
\author[label1]{Jiabo Ye}
\ead{52194506006@stu.ecnu.edu.cn}
\author[label1]{Linlin Wang \corref{mycorrespondingauthor}}
\cortext[mycorrespondingauthor]{Corresponding author}
\ead{llwang@cs.ecnu.edu.cn}
\author[label1]{Xin Lin}
\ead{xlin@cs.ecnu.edu.cn}
\author[label1]{Liang He}
\ead{lhe@cs.ecnu.edu.cn}
\address[label1]{The Department of Computer Science and Technology, East China Normal University, Shanghai 200241, China.}

\begin{abstract}
Inferring the substitutable and complementary products for a given product is an essential and fundamental concern for the recommender system.
To achieve this, existing approaches take advantage of the knowledge graphs to learn more evidences for inference, whereas they often suffer from invalid reasoning for lack of elegant decision making strategies. Therefore, we propose a novel Knowledge-Aware Path Reasoning (KAPR) model which leverages the dynamic policy network to make explicit reasoning over knowledge graphs, for inferring the substitutable and complementary relationships. Our contributions can be highlighted as three aspects. 
Firstly, we model this inference scenario as a Markov Decision Process in order to accomplish a knowledge-aware path reasoning over knowledge graphs.
Secondly, we integrate both structured and unstructured knowledge to provide adequate evidences for making accurate decision-making.
Thirdly,  we evaluate our model on a series of real-world datasets, achieving competitive performance compared with state-of-the-art approaches.
Our code is released on https://gitee.com/yangzijing\_flower/kapr/tree/master
.

\end{abstract}


\end{frontmatter}

\section{Introduction}
Understanding substitutable and complementary relationships between products boosts the development of recommender system \cite{tintarev2007survey_UserSatisfaction,zhang2016collaborative_KGEmbedding,huang2018improving_RS,newadd1,newadd2}.
Substitutable relationship links two products with similar functions and can be substituted with each other. 
Complementary relationship links two products with same usage scenarios but playing complementary roles. 
These two product relationships are of great significance when recommended to users with different purchase intentions \cite{wu2017session_PurchaseIntension}.
For instance, it is reasonable to recommend a Samsung Galaxy S10 Plus as a substitutable product for a user who is browsing the page of iPhone11 Pro.
Meanwhile, after the purchase, the recommender system should recommend phone's complements, for instance, USB charging cable, to the user.
The contribution of this article to the community can be viewed from both the user and the e-commerce platforms. From the user's point of view, the significance of the research is to make recommendations based on the user's behavior, make recommendations that are more in line with the user's preference, improve the user's shopping efficiency. From the perspective of e-commerce platforms, the significance of this task is to reduce redundant recommendations and increase the success rate of transactions. 

The mainstream of existing researches focus on modeling product representation, which can be categorized into text-based methods and relation-based methods. The text-based approaches use different methods to learn product representation, such as Latent Dirichlet Allocation (LDA) \cite{blei2003latentLDA} and variational autoencoders (VAE) \cite{rakesh2019LVAE}. The relation-based methods learn product representation through the relationship constraints between products, such as PMSC \cite{DBLP:conf/wsdm/WangJRTY18_JD} and SPEM \cite{zhang2019inferring_SPEM}. They incorporate product embedding with constraints to further promote the distinction between two relationships. DecGCN encodes the semantic representation of products through GCN on an product graph and model product substitutability and complementarity in two spaces \cite{cikm2020}.

Although the previous methods have made progress, the exploration of topological information of knowledge graph is still under-exploited. Most of the existing methods suffer from two shortcomings.
(1) None of these models considers the issue of fine-grained inference, which takes into account product characteristics. Judgments solely rely on product representation fails to derive the fine-grained characteristics of their relevance. For example, they can't find a substitute featuring `Bluetooth' to a given keyboard.
(2) Inferences lack interpretability. Existing models can only estimate the probabilities of the substitutable and complementary relationships, but the reason cannot be further explained. The inferring result may not be convincing if the algorithm can not be interpreted.
\begin{figure}
  \centering
  \includegraphics[width=1\textwidth]{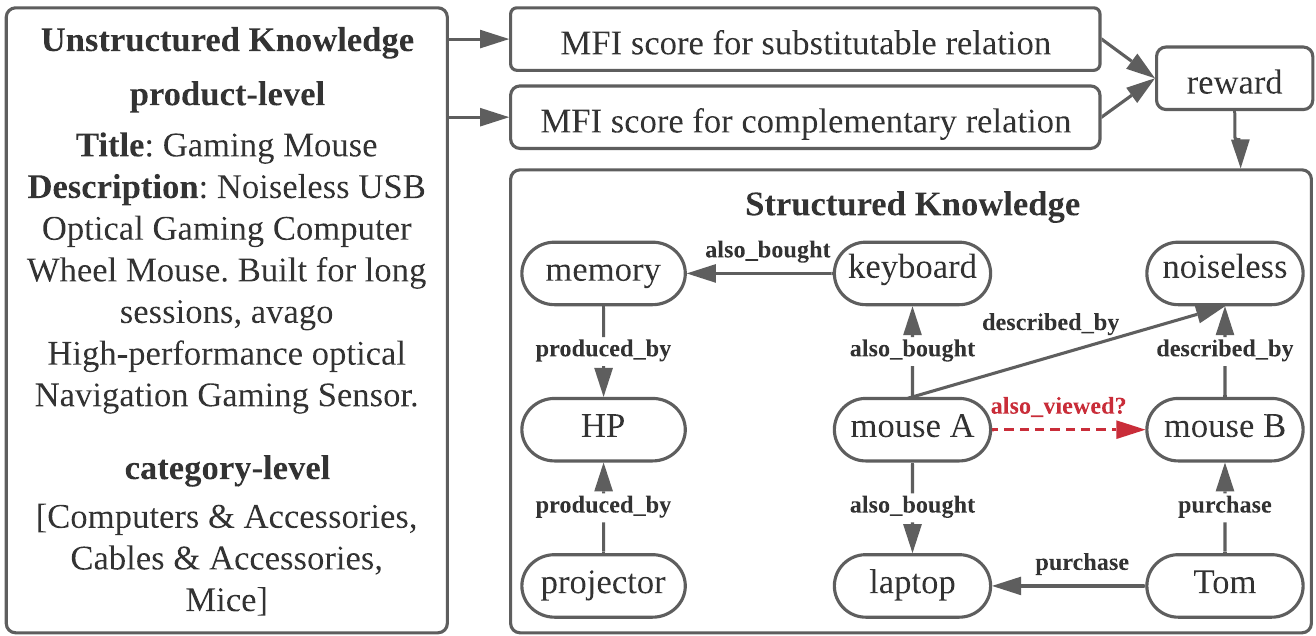}
  \caption{Illustration of path reasoning in KAPR. The dashed line indicates whether the relationship exists between two products.}
  \label{Figure 1}
\end{figure}

To overcome these two shortcomings, we propose a Knowledge-Aware Path Reasoning (KAPR) method to infer substitutable and complementary relationships over knowledge graph. Specifically, we first construct a knowledge graph by extracting structured information from e-commerce datasets to better tackle  the problems in fine-grained inference and interpretability \cite{he2016ups_dataset}. Then we design different meta-paths comprising of fine-grained characteristics (e.g., brand, word, category) for modeling substitutable and complementary relationships, which enables the agent to navigate over potential products with specified characteristics through these meta-paths. Based on these, the proposed model can cast the product relationship inference task into a path reasoning problem over knowledge graph and make accurate decisions with dynamic policy networks. 

The superior advantage of KAPR is that we elaborately incorporate both structured and unstructured knowledge to explicitly guide the reasoning path and increase the interpretability of  the model. On one hand, we leverage structured knowledge to construct a graph with constraints to provide the model with necessary adjacent information and eliminate many irrelevant products, effectively narrowing the search space with a pruning strategy to ensure reasoning accuracy. On the other, we propose a Multi-Feature Inference (MFI) component  based on relevant unstructured knowledge (e.g. title, description, category, and  relevance) to construct a novel reward function in order to guide the learning process.  

An example is illustrated in Figure\ref{Figure 1}. Given an entity `mouse A', we expect the proposed model to find a candidate product which has a substitutable relationship with `mouse A'.  To perform this inference, the KAPR first constructs a knowledge graph shown in the lower right part of the figure, and then utilizes a dynamic policy network to 
reason over this graph. More precisely, starting from `mouse A', the agent first obtains adequate features from adjacent entities to search for possible paths with two-hop traversal over knowledge graph, and then adopt the pruning strategy to narrow  search space into a valid subspace, including two paths `mouse A — noiseless — mouse B' and `mouse A — keyboard — memory'. Next, this agent relies on MFI component to exploit unstructured knowledge from both product and category-level  (e.g. `Computers \& Accessories') to obtain relevance scores of substitutable products, constituting  reward function to guide path reasoning toward the right direction. Finally, the KAPR  reaches  the target entity `mouse B'  with a highest rank. 

Significant contributions of this work can be summarized as follows:
\begin{itemize}
  \item[1.]  To accurately infer the substitutable and complementary relationships, we propose a Knowledge-Aware Path Reasoning (KAPR) model, which leverages dynamic policy networks, to perform in-depth reasoning over knowledge graphs, opening up the avenue of utilizing knowledge-aware reasoning into fine-grained product relationship inference. 
  \item[2.] We integrate both structured features from knowledge graph and unstructured textual information to provide adequate evidences for exact decision-making. Meanwhile, the reward function newly constituted from unstructured knowledge also enables the path reasoning toward a right direction over knowledge graph. 
  \item[3.]  The extensive experimental evaluation on real-world datasets shows that the our model can successfully tackle substitutable and complementary relationship inference with interpretable path reasoning, surpassing all previous state-of-the-art approaches.
\end{itemize}
\section{Related Work}
\subsection{Product Relationship Inference}
In this paper, we mainly focus on substitutable and complementary relationships  between products in recommender systems. 
Previous studies formulate the task of substitutable and complementary relationship inference as a supervised link prediction problem. Sceptre is the first proposed model to tackle the inference which constructs product representation from textual information through LDA \cite{mcauley2015inferringSceptre}.
However, Sceptre neglects the relationships information and does not always work well since LDA is not effective for short texts. Further, PMSC adopts a novel loss function with relation constraints to distinguish between the substitutes and complements\cite{DBLP:conf/wsdm/WangJRTY18_JD}, and LVAE links two variational auto-encoders to learn  latent features over product reviews\cite{rakesh2019LVAE}. SPEM considers both textual information and relational constraints \cite{zhang2019inferring_SPEM}, and DecGCN exploits the graph structure to learn product  representations in different relationship spaces \cite{cikm2020}. Nevertheless, all these methods suffer from two drawbacks. Firstly, these models can not perform a fine-grained inference since the embedding modules donot take word-level characteristics into consideration. Secondly, these models are lack of explainability due to the fact that probability values cannot reveal the original relationships. Therefore, we aim to propose a novel framework  to provide accurate recommendations with explicit inferences.
\subsection{Reinforcement Learning in Recommendation}
Recently, many reinforcement learning (RL) based models applied in recommender systems \cite{chen2019generative,xiong2017deeppath_FormulationReasoning}, such as multi-agent RL-based model \cite{gui2019mention}, the hierarchical RL-based model \cite{zhang2019hierarchical}.
Reinforcement learning has received a series of attention, it can understand the environment, and has a certain reasoning ability. Reinforcement learning has been widely used in recommender systems, such as product recommendation \cite{xian2019reinforcement}, advertisement recommendation \cite{DRN} and explainable recommendation \cite{ER}. Among them, Xian regards users, commodities and their related attributes as nodes, and trains the agent to find the potential purchase relationship between user and products[1]. Based on the Q-Learning method, Zheng et al. simulate the rewards given by users to complete the news recommendation task[2]. Wang et al. designed an interpretability framework based on reinforcement learning, which can be flexibly explained according to usage scenarios[3].
 PGPR \cite{xian2019reinforcement} is an RL-based path reasoning model for personalized recommendation. Compared with previous RL-based models, PGPR achieves higher accuracy and can give 
explicit evidences for inference.
However, its policy network is static, unable to encode large-scale action spaces. PGPR only uses structured knowledge for reward function, ignoring the information in unstructured knowledge, resulting in poorly guided and unwell reward signals.
In this paper, our method addresses the issues by proposing a knowledge-aware path reasoning method and integrating structured and unstructured knowledge to guide the reasoning.

\section{Problem Formulation}
This task aims to find the substitutable and complementary products for a given product $v_{0}$.
We define the knowledge graph as $G=\{(e,r,e^{'})|e,e^{'} \in E, r \in R\}$, where $E$ is the entity set and $R$ is the relation set which consists of six type of relationships.
$e$ and $e^{'}$ represent two different types of entities, and $r$ represents the relationship between entities.

The relationship tuples are `product-described\_by-word', `product-belong-category', `product-produced\_by-brand', `product-purchase-user', `product-also\_bought-product', `product-also\_viewed-product'.
The tuple $(e,r,e^{'})$ reflects the facts that $e$ and $e^{'}$ have a relationship $r$, where $e(e^{'})$ is a certain entity and $r$ is a type of relationship. 

`Also\_bought’ and ‘also\_viewed’ are abstracts from user behavior. `Also\_bought’ indicates that two products are often purchased together. `Also\_viewed’ indicates that two products are often viewed together. We follow the definition of previous work, and call two products with also\_bought relationship as complementary products, and two products with also\_viewed relationship as substitutable products.
In this setting, substitutable product inference requires the model to find a product set $\hat{V}_{v_{o}} \subset V$ that has a relationship of ‘also viewed’ with the starting product $v_{0}$ with explicit reasoning path $P_{v_{0},v_{i}}$ consists of the tuples, whereas complementary product inference requires the model to find certain products which have a relationship of ‘also\_bought’.
\section{The Proposed Model}
\begin{figure*}
\centering
\includegraphics[width=1\textwidth]{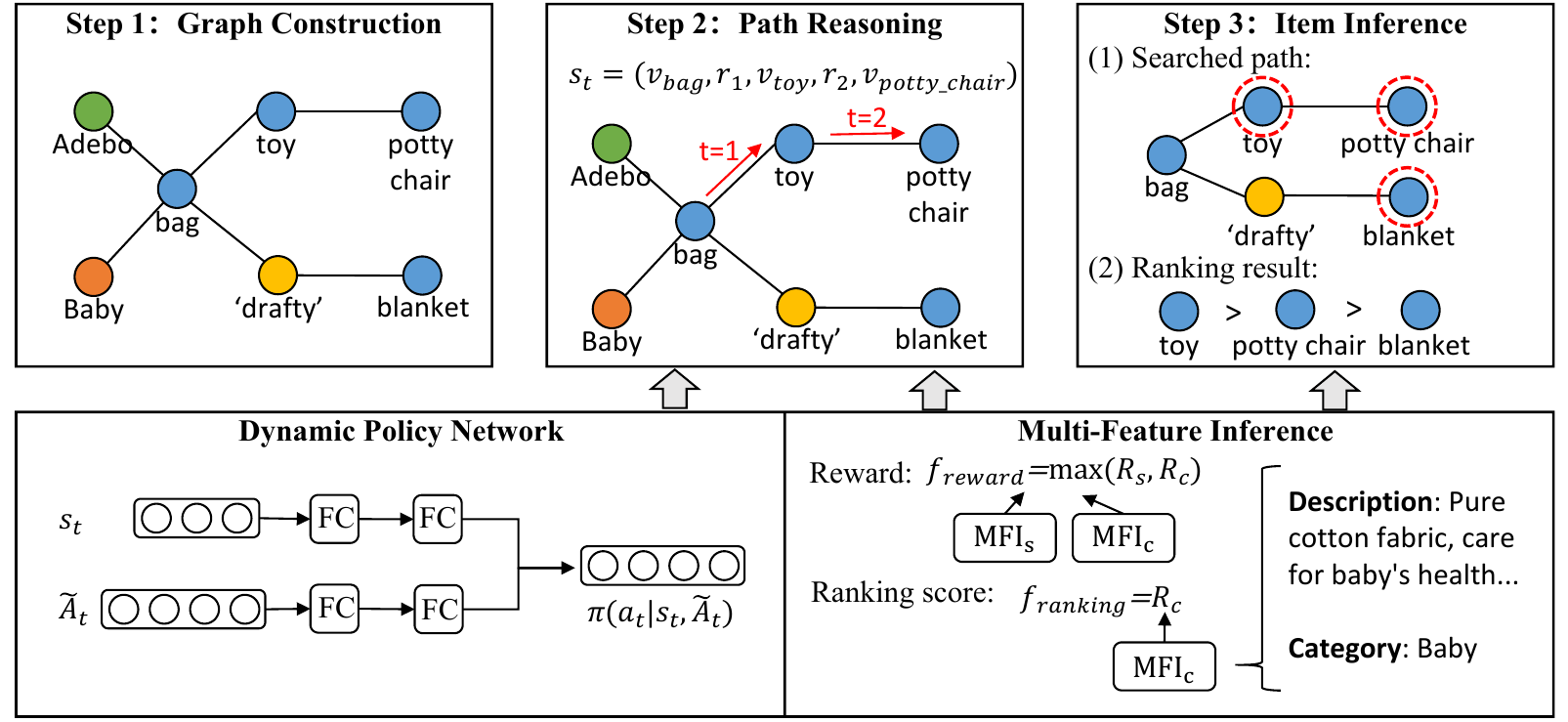}
\caption{The pipeline of Knowledge-Aware Path Reasoning framework. 
The overall process is divided into three main stpdf: graph construction, path reasoning, and product inference.
The gray arrows represent the relevant component that participates in the specific process.
The first step is to construct a graph based on the attributes of the product. The nodes in blue stand for products, green nodes refer to brands, red nodes represent categories, and yellow nodes are the words in reviews.
At the path reasoning stage, the agent interacts with the environment to find the correct reasoning path (the red arrows) with the dynamic policy network and multi-feature inference. This policy enables the agent to navigate from a given product (bag) to the desired target product (potty chair) based on the adjacent attributes in knowledge graphs.
As for item inference, we sort the products by ranking MFI\_s or MFI\_c (the scores of multi-feature inference) to obtain the list of the complement or substitute. All products on the paths are candidates marked with dotted circle borders.}
\label{Figure 2}
\end{figure*}

In this section, we propose the Knowledge-Aware Path Reasoning method (KAPR) to infer substitutable and complementary relationships over knowledge graph. The overall structure of the proposed model is illustrated in Figure\ref{Figure 2}. In this method, we formulate this relationship inference task as an MDP environment and adopt an agent to navigate the products with a potential relationship (substitutable or complementary) of the given product. 
At the inference stage, the trained agent samples a series of paths for each product, and all  products on the path constitute the candidate product sets. The final ranking of substitutable and complementary products are obtained by sorting the candidate product sets based on corresponding scores. 

The rest of this section is organized as follows: we will start with the formulation of MDP in Section~\ref{mdp}, and then introduce detailed working principles of knowledge-aware path reasoning in Section~\ref{reasoning}. In Section~\ref{inference}, we will describe the final step which conducts  the relationship inference with knowledge-aware  reasoning via dynamic policy network on the graph.

\subsection{Formulation as Markov Decision Process}
\label{mdp}
The knowledge graph $G$ has a relation set $R$ and an entity set $E$ which consists of a product set $V$, a word set $W$, a brand set $B$, a category set $C$ and a user set $U$. The definition of entities and relationships is in Table\ref{Table 1}. 
Therefore, the graph can be represented as $G=\{(e,r,e^{'})|e,e^{'} \in E, r \in R\}$, where each tuple stands for the fact that $e$ and $e^{'}$ have a relationship $r$ belonging to the above six types. 
We regard each $(e,r,e^{'})$ as a tuple, representing a triplet relationship in the knowledge graph $G$.
There are 6 kinds of tuples in the knowledge graph, representing 6 kinds of relations between entities in the e-market scene
The tuples have the same meaning as previous work \cite{xian2019reinforcement}. 
$v$, $b$, $c$, $u$, $w$ represent products, brands, categories, users, and words respectively, where $v \in V$, $b \in B$, $c \in C$, $u \in U$, $w \in W$, $v_{i} \in V$ and $0 \leq i \textless |V|$.
The types of tuples include ($v_{1}$, also\_viewed, $v_{2}$), ($v_{1}$, also\_bought, $v_{2}$), ($v$, belong to, $c$), ($v$, produced by, $b$), ($v$, purchase, $u$), ($v$, described, $w$).

We follow the terminologies commonly used in reinforcement learning to describe the MDP environment \cite{shani2005_mdp}.
The environment informs the agent search state $s_{t}$ and the complete action apace $A_{t}$ at time $t$.
When the agent finds a product, it gets the reward $R$. 
Formally, the MDP environment can be defined by a tuple $(S, A, \delta, \rho)$, where $S$ denotes the state space, $\delta:S \times A \rightarrow S$ refers to the state transition function, and $\rho: S \times A \rightarrow R$ is the reward function. We start the path reasoning process at product $v_{0}$.
\begin{itemize}
\item \emph{State}: The initial state $s_{0}$ is represented as $s_{0}=v_{0}$. When we consider $k$-step history, the state $s_{t}$ at step $t$ is defined as Equation \ref{eq:state}.
\begin{equation}\label{eq:state}
    \begin{aligned}
   s_{t}=(v_{0},r_{t-k},e_{t-k},...,r_{t-1},e_{t-1},r_{t},e_{t}).
\end{aligned}
\end{equation}
\item \emph{Action}: For state $s_{t}$, the action is defined as $a_{t}=(r_{t+1}, e_{t+1})$, where $e_{t+1}$ is the next entity and $r_{t+1}$ is the relationship that connects $e_{t}$ and $e_{t+1}$. The complete action space $A_{t}$ is defined as all edges connected to entity $e_{t}$ excluding history entities and relationships. 
Because some nodes have a very large outgoing degree, we propose a pruning strategy with structured knowledge for action pruning.
The pruned action space is represented as $\tilde{A}_{t}$. The detail of the pruning strategy is further explained in the next section.
\item \emph{Transition}:
Given state $s_{t}$ and action $a_{t}$, the transition to the next state $s_{t+1}$ is:
\begin{equation}\label{eq:transition}
    \begin{aligned}
    s_{t+1}&=P(s_{t},a_{t}) \\
    &=(v_{0},r_{t-k},e_{t-k},...,r_{t-1},e_{t-1},r_{t},e_{t},r_{t+1},e_{t+1})\\
\end{aligned}
\end{equation}

\item \emph{Reward}: 
In the path finding process, for the model to learn the fundamental reasoning process, we design different meta paths for the two relationships. Only when the agent generates a path that fits in the meta paths and ends with a product, a reward is calculated. In other cases, the reward is 0. For the relationship sparseness problem, using binary rewards can lead to inadequate supervision. Instead, we use a reward function with unstructured Knowledge to guide the reasoning. The detail of the reward function is further explained in the next section.
\end{itemize}
\subsection{Knowledge-Aware Path Reasoning}
\label{reasoning}
Now we present the pruning strategy with structured knowledge, dynamic policy network and the reward function with unstructured knowledge .

\subsubsection{Pruning Strategy with Structured Knowledge}
Since some nodes have much larger out-degrees, we propose a pruning strategy using structured knowledge.
This strategy needs to retain actions that help inference keeping nodes that are closely associated with $v_{0}$.
The pruning strategy consists of two stpdf. 
First, we exclude impossible edges based on the meta path patterns, then the scoring function $f((r, e) | v_{0})$ maps all actions $(r, e)$ to a value conditioned on the starting product $v_{0}$. We regard the top-n actions as the pruned action space of the state $s_{t}$.
In this paper, we use TransE to initialize all entity and relationship representations \cite{bordes2013translating_TransE}.
All types of entities have a 1-hop pattern with the product entity, such as $\text{user}\stackrel{\text{purchase}}{\longleftrightarrow}\text{product}$, $\text{product}\stackrel{\text{also\_viewed}}{\longleftrightarrow}\text{product}$.
There are two relationships (substitutable and complementary) between product entities, we choose the maximum of the two scores. The scoring function is defined as Equation \ref{eq:score function00}.

\begin{equation}
\label{eq:score function00}
    f((r,e) | v_{0})=
\begin{cases}
<\mathbf {v_{0}}+\mathbf r_{d},\mathbf e>+b_{e}\ ,\ e \notin V\\
\text{max}(<\mathbf v_{0}+\mathbf r_{also\_viewed},\mathbf e>+b_{e},\\<\mathbf v_{0}+\mathbf r_{also\_bought},\mathbf e>+b_{e})\ ,\ e \in V
\end{cases}
\end{equation}
Here, $r_{d}$ is the relationship that directly connects product nodes and other types of nodes.

\subsubsection{Dynamic Policy Network}
Based on the MDP formulation, our goal is to learn a policy network $\pi$ that maximizes the expected cumulative reward for the path reasoning.
We design a dynamic policy network that can select actions in a dynamically changing space. 
The policy network $\pi(\cdot|s,\tilde{A}_{v_{0}}))$ takes the state embedding $\mathbf s$ and action embeddings $\mathbf a_{s}$ as input and emits the probability of each action. 

We map $\mathbf s$ and $\mathbf a_{s}$ into a shared learnable feature space and compute their affinity between $\mathbf s$ and each action and we apply a softmax function to normalize the affinity into a probability distribution. The structure of the dynamic policy network is defined in Equation \ref{eq:dynamic}.
\begin{equation}
\label{eq:dynamic}
    \begin{aligned}
    &\mathbf s^{'}=\text{ReLU}(\text{ReLU}(\mathbf s \mathbf W_{s})\mathbf W_{1}) \\
    &\mathbf a_{s}=\tilde{A}_{v_{0}}\mathbf W_{A} \\
    &\mathbf a^{'}_{s}=\text{ReLU}(\text{ReLU}(\mathbf a_{s}\mathbf W_{a})\mathbf W_{2}) \\
    &\pi(\cdot|s,\tilde{A}_{v_{0}})=\text{softmax}(\mathbf s^{'} \otimes \mathbf a^{'}_{s})\\
    \end{aligned}
\end{equation}
Here, $\mathbf s$ and $\mathbf s^{'}$ stand for the embedding and hidden features of the state,$\mathbf W_{A}$ is an action-to-vector lookup table, $\tilde{A}_{v_{0}}$ stands for the pruned action space.
$\mathbf a$ and $\mathbf a^{'}$ are the embedding and hidden features of all actions in $\tilde{A}_{v_{0}}$.
$\mathbf s \in R^{d_{s}}$, $\mathbf s^{'}\in R^{d_{s’}}$, $\mathbf a_{s} \in R^{d_{a} \times M}$. $\mathbf a^{'}_{s} \in R^{d_{a’} \times M}$, $M \leq D$, $M$ is the size of the space action and $D$ is the maximum size of the space action.
$\mathbf s$ is represented as the concatenation of the embedding $(\mathbf v_{0},\mathbf r_{t-k},\mathbf e_{t-k},...,\mathbf r_{t-1},\mathbf e_{t-1},\mathbf r_{t},\mathbf e_{t})$.
Each action $\mathbf a_{i}$ is represented as the concatenation of the embedding $(\mathbf r_{t+1},\mathbf e_{t+1})$. 
$\mathbf a_{s}$ is the concatenation of the embedding ($\mathbf a_{0},..., \mathbf a_{M}$).
$\mathbf v,\mathbf e_{t},\mathbf r_{t}$ are product, entity and relationship embeddings learned by TransE.
The model parameters for both networks are denoted as $\theta=\{\mathbf W_{1},\mathbf W_{2},\mathbf W_{s},\mathbf W_{a}\}$. The policy network gradient  is defined as Equation \ref{loss}.
\begin{equation}
\label{loss}
\begin{aligned}
\nabla_{\theta}J(\theta)=E_{\pi}[{\nabla_{\theta}}log \pi_{\theta}(\cdot|s,\tilde{A}_{v_{0}})G]
\end{aligned}
\end{equation}
Here, $G$ is the discounted cumulative reward from the initial state to the terminal state.
\subsubsection{Reward Function with Unstructured Knowledge}
\begin{figure}
  \centering
  \includegraphics[width=1\textwidth]{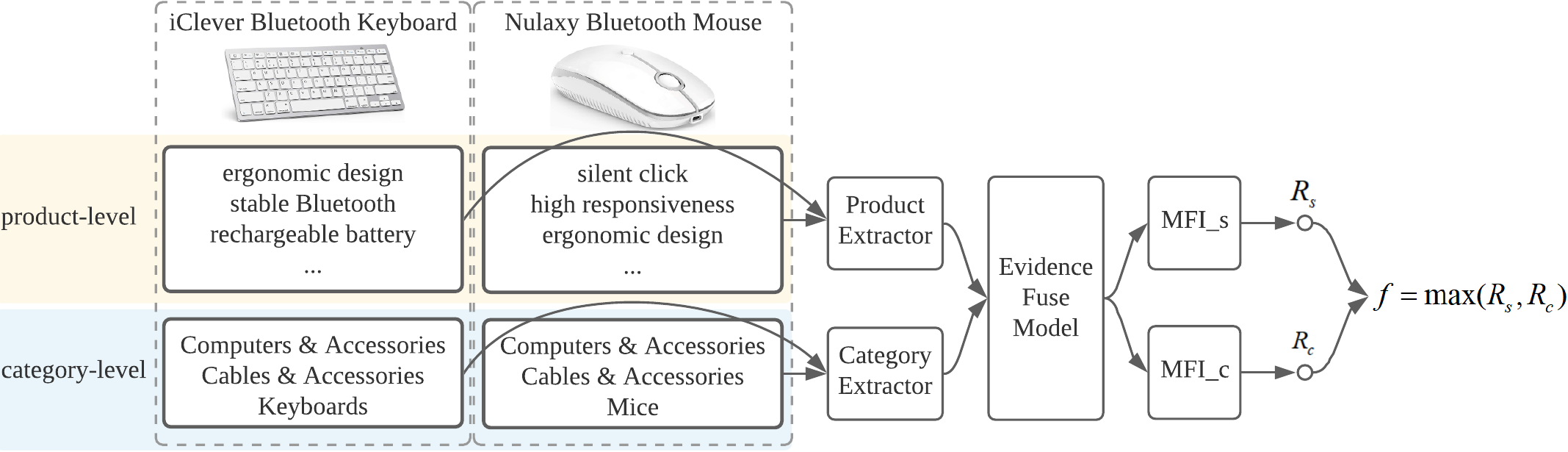}
  \caption{The overall framework of the reward function model. The reward function model consists of two $\text{MFI}$ model. The model for predicting the substitutable relationship named MFI\_s and the model for predicting the complementary relationship called MFI\_c. The reward of products is the maximum value of MFI\_s and MFI\_c, which represents the correlation of two products.}
  \label{Figure 3}
\end{figure}

The training of the policy network needs rewards that measure the relevance between two products.
A general method uses TransE to calculate the distance between the entities as reward \cite{xian2019reinforcement}, but this method is not applicable in the sparse graph.
Substitutable and complementary relationship is very sparse. The sparse reward leads to convergence problems in the path reasoning process. To solve this problem, we propose a model for extracting unstructured knowledge to generate more robust rewards. Unstructured knowledge contains more information, such as textual knowledge (product's reviews and descriptions), which can accurately reflect the relevance of two products. 
We propose a Multi-Feature Inference component (MFI) as a prediction model  that infers relationship based on category-level and product-level features.
Introducing category-level features can provide shared semantic information among products that belong to the same category. 
Although labelled product pairs are sparse compared with the overall product set, MFI can learn the shared knowledge on category-level with product-level supervision signal.
Figure\ref{Figure 3} shows an overview of the reward function model which consists of two MFI models.
We consider the title of all products under the category as the document due to they usually give the main topic. First, we extract the most important $F$ words in each document through TF-IDF \cite{zhang2017joint_TFIDF}. For category $c$ , we regard its textual knowledge as $T=[w_{1}, w_{2},..., w_{N}]$. We input these words into embedding through Glove vector \cite{pennington2014glove} and form an embedding sequence as $\mathbf T=[\mathbf w_{1}, \mathbf w_{2},..., \mathbf w_{N}]$ , $\mathbf w_{i} \in R^{d_{c}}$, where $d_{c}$ is the dimension of word embedding.

For product-level features, we firstly collect the descriptions of all products.
Then a doc2vec\footnote{https://radimrehurek.com/gensim/models/doc2vec.html} model is trained by the corpus of each category \cite{le2014distributed_doc2vec}. The model is used to infer a vector $\mathbf v_{i}$, $\mathbf v_{i} \in R^{d_{p}}$ for the textual information of each product $i$. Product-level features are extracted from the well-trained model, and it is feasible to directly use them as product representation. However, through our experiment, it can achieve better results by applying a module consists of an MLP with mask-attention and BatchNormal \cite{DBLP:conf/nips/VaswaniSPUJGKP17Attention}. 
The mask-attention reads a feature vector and generates a soft mask vector with the same shape, and the mask-attention $F_{attn}$ is modeled by  Equation \ref{eq:attn}.
\begin{equation}\label{eq:attn}
    \begin{aligned}
    &a=\sigma\left(W \mathbf v_{i}+b\right) \\
    &\mathbf v_{i}^{'}=a\odot \mathbf v_{i}\\ 
\end{aligned}
\end{equation}
The multi-layer perception consists of one mask-attention $F_{attn}$ and $L$ linear layers $F_{mlp}$. We can formulate the entire muti-layer perception as Equation \ref{evid_prod}.
\begin{equation}\label{evid_prod}
\begin{aligned}
    &y^{\left(0 \right)}=F_{attn}\left(\mathbf v_{i}  ;\theta_{attn} \right)\\
    &y^{\left(k \right)}=F_{mlp}\left(y^{\left(k-1 \right)} ;\theta_{mlp}^{\left(k \right)}\right),k=1,...,L
\end{aligned}
\end{equation}
Here, the ﬁnal result $y^{L} \in R^{d_{p}}$ is the product-level evidences. 

For link prediction, MFI emits the probability that $v_{i}$ and $v_{j}$ belong to relationship $r$. We use a nonlinear classifier to project all evidences to the probability. The loss function is defined as Equation \ref{MFI}.

\begin{equation}\label{MFI}
\begin{aligned}
\mathcal{L}=&\sum_{i,j \in \varepsilon_{P}}{y_{i,j} \cdot \log \text{MFI}\left(i,j\right)}+\\
&\sum_{i,j \in \overline{\varepsilon}_{P}}{\left(1-y_{i,j}\right) \cdot \log \left(1-\text{MFI}\left(i,j\right)\right)}
\end{aligned}
\end{equation}

The models to predict the substitutable and complementary relationships are named as MFI\_s and MFI\_c, respectively.
In our experiments, we find that regardless of target relationship, substitutable and complementary relationship play an essential role in the reasoning process. 
For instance, two chargers substitute to each other and both of them are complements to a phone. 
No matter what kind of relationship the agent finds, both substitutes and complements are beneficial to the reasoning. 
So we use the maximum value of MFI\_s and MFI\_c as a reward.
The reward function is defined as Equation \ref{MFISC}.
\begin{equation}\label{MFISC}
\begin{aligned}
f(v_{0},e_{t})=\text{max}(\text{MFI}\_\text{s}(v_{0},e_{t}),\text{MFI}\_\text{c}(v_{0},e_{t}))
\end{aligned}
\end{equation}
Here, $\text{MFI}\_\text{s}(v_{i},v_{j})$ and $\text{MFI}\_\text{c}(v_{i},e_{j})$ represent the probability that $v_{i}$ and $v_{j}$ have a substitutable or complementary relationship.
\subsection{Inference Techniques}
\label{inference}
The final step is to solve the product relationship inference problem with knowledge-aware path reasoning via policy network on the graph. 
Our goal is to find product set $\{v_{i}\}$ that has a relationship $r$ with starting product $v_{0}$ and the explicit reasoning path $P_{v_{0},v_{i}}$ for interpretability.
To ensure the diversity of paths, the agent should not repeatedly search for paths with big rewards. 
Therefore, we employ beam search to select actions guided by the dynamic policy network.
Given a product $v_{0}$, dynamic policy network $\pi(\cdot|s,\tilde{A}_{v_{0}}))$, horizon $T$ and sampling size at each step $t$, represented by $K_{1},\cdots,K_{T}$. The search path at step $t$ is $SP_{t}$, $SP_{0}=v_{0}$.

The reasoning process at step $t$ is as follows.
First, the agent from the last retrieved node in $SP_{t}$ and obtains the pruned action space according to the structured knowledge the meta-path patterns.
The meta-path patterns allow the agent to learn basic reasoning modes and its search direction is determined by structured knowledge.
Secondly, employing beam search based on the action probability emitted by the dynamic policy network, the agent selects $K_{t}$ actions with the highest probability to search.
The dynamic policy network can handle dynamically changing action spaces and be rewarded by unstructured knowledge which means it can choose appropriate actions according to the current state.
Thirdly, Save the retrieved path to $SP_{t+1}$.

After $T$ stpdf of search, we get a network composed of paths. All nodes in the network are closely related to product $v_{0}$, and the connection path between them can be used as explicit evidence of the association. 
We save paths with length greater than 1 and end with a product.
These nodes may be substitutes or complements of $v_{0}$. Then we score these paths according to MFI\_s or MFI\_c, aiming to distinguish substitutes and complements.
Finally, the products that appear in the training set are removed and we select top-N as the final inference result.

\section{Experiments}
We evaluate our model on four datasets to demonstrate the improvements. 
We firstly introduce the statistics of datasets, evaluation metrics, and experimental settings. 
And then, we present the statistics of all model’s performances and provide an ablation study to investigate the effects of each component of the model.
Next, we conduct a series of detailed analyses to illustrate the effectiveness and interpretability of our model.
Finally, we make an error analysis of the experimental results.
\subsection{Datasets and Evaluation Metrics}
\subsubsection{Dataset}
We evaluate our model with 4 datasets \cite{he2016ups_dataset}: Baby, Beauty, Cell Phone (Cell Phones and Accessories) and Electronics. Each dataset contains reviews and product metadata. 
The definition and statistics of entities and relations in each dataset can be found in Table\ref{Table 1}. We follow the method in previous work to keep the critical words in the reviews as features \cite{xian2019reinforcement}.
We randomly choose 85\% of the data for training and the remain for testing. When training the MFI model, we randomly sample $N$ non-links from the dataset. When training the KAPR model,  we treat all products as candidate products instead of negative sampling.

\begin{table}[]
\centering
\footnotesize
\caption{The definition and statistics of each dataset.}
\begin{tabular}{c|cc|cccc}
\hline
\multicolumn{3}{c|}{}                                                     & Baby       & Beauty     & Cell Phone     & Electronics    \\ \hline
Entity        & \multicolumn{2}{c|}{Definition}                           & \multicolumn{4}{c}{Number of Entities}                    \\ \hline
Product       & \multicolumn{2}{c|}{Products in the dataset}              & 10027      & 20019      & 10018          & 10023          \\
User          & \multicolumn{2}{c|}{Users in the dataset}                 & 18990      & 20198      & 21880          & 157283         \\
Word          & \multicolumn{2}{c|}{Words in reviews} & 16909      & 27975      & 14850          & 19808          \\
Brand         & \multicolumn{2}{c|}{Brand of products}    & 1062       & 2615       & 731            & 1302           \\
Category      & \multicolumn{2}{c|}{Categories of Products}               & 1          & 237        & 104            & 607            \\ \hline
Relation      & Head Entity                  & Tail Entity                & \multicolumn{4}{c}{Number of Relations per Head Entities} \\ \hline
Also\_viewed  & Product                      & Product                    & 27.144     & 22.402     & 16.647         & 14.474         \\
Also\_bought  & Product                      & Product                    & 33.796     & 21.005     & 28.473         & 29.682         \\
Described\_by & Product                      & Word                       & 12.834     & 14.437     & 14.562         & 7.165          \\
Produced\_by  & Product                      & Brand                      & 0.756      & 0.757      & 0.612          & 0.686          \\
Belong\_to    & Product                      & Category                   & 1.000      & 4.141      & 3.265          & 4.390          \\
Purchase      & User                         & Product                    & 9.158      & 3.620      & 5.202          & 44.441         \\ \hline
\end{tabular}
\label{Table 1}
\end{table}

\subsubsection{Evalution Metrics}
To measure the inference accuracy, we adopt the evaluation method in SPEM, Hits@k to evaluate rank-based methods.
We vary the value of $k$ by \{10, 30, 50\}. For each product pair ($A$, $B$) in the test set, we take the evaluation stpdf for Hits@k as follow \cite{zhang2019inferring_SPEM}.
(1) We randomly sample $n$ products with which product $A$ is irrelevant, $n$ is set to 500. 
(2) Among the $n$ products, the number of products ranking before product $B$ is $m$. 
(3) If $m<k$, we get a hit. Otherwise, we get a miss. 

To further compare with the path reasoning model PGPR, we use Normalized Discounted Cumulation Gain (NDCG), Hit Ratio (HR), Recall and Precision as the evaluation metrics.
We evaluate the metrics based on the top 10 recommended products for each product in the test set. 
For product $A$, we regard all products in the dataset as candidate products, except for products in the training set. we take stpdf as follow:
(1) We get the candidate products list $L$ based on MFI score.
(2) We evaluate the metrics based on the top-$k$ recommended products. $k$ is set to 10.
\subsection{Experimental Settings}
The parameter setting for the MFI model is as follow: 
$F$ is set to 15 for TF-IDF, that is, for each review, the 15 most critical words are selected as representatives.
The dimension of each feature are $d_p=300$ and $d_c=100$. The hyper-parameter values for doc2vec: vector size=300. window size=20. 
For the MDP  environment, we mainly refer to PGPR.
The history length $K$ = 1 for the state $s_{t}$ and the maximum length $T$ = 3 for the reasoning.
The sampling size is [25, 5, 1].
We set the maximum size of the action space $D$ = 250 and the embedding size of entities and relationships is $d_{e}$ = 100.
The learning rate for our model is 0.001, and the batch size is 16.

\subsection{Results and Analysis}
We compare KAPR with the following models in substitute and complement product inference. 
Sceptre uses LDA to extract features from product textual information to predict the relationships \cite{mcauley2015inferringSceptre}. 
LVA links two VAE to learn the content feature of products \cite{rakesh2019LVAE}.
SPEM applies a semi-supervised deep Autoencoder to preserve the second-order proximity between products \cite{zhang2019inferring_SPEM}. SPEM can only predict the substitute. 
PGPR is an RL-based path reasoning model for personalized recommendation \cite{xian2019reinforcement}. It adopts a policy-based method to reason in the knowledge graph. We adjust the model so that PGPR can infer the product relationship. 
DecGCN models the substitutability and complementarity of products in separated embedding spaces \cite{DecGCN}.
To test the performance of the MFI model, we treat MFI as a supervised prediction model to infer relationship.

\begin{table}[]
\centering
\small
\caption{Hits@k of all methods on four datasets for the substitutable and complementary relationship.}
\begin{tabular}{c|c|ccc|ccc}
\hline
                             & Relation & \multicolumn{3}{c|}{Substitute}                  & \multicolumn{3}{c}{Complement}                  \\ \hline
Category                     & Method   & K@10           & K@30           & K@50           & K@10          & K@50           & K@50           \\ \hline
\multirow{6}{*}{Baby}        & LVA      & 0.47           & 0.58           & 0.63           & 0.28          & 0.36           & 0.40           \\
                             & Sceptre  & 0.25           & 0.44           & 0.86           & 0.36          & 0.55           & 0.62           \\
                             & SPEM     & 0.89           & 0.90           & 0.92           & --             & --              & --              \\
                             & PGPR     & 0.66          & 0.97          & 0.98          & 0.47         & 0.81          & 0.88          \\
                             & DecGCN   & 0.74           & 0.98           & 0.98          & 0.52         & 0.83           &0.89            \\
                             & MFI      & 0.55           & 0.75           & 0.84           & 0.25          & 0.42           & 0.52           \\
                             & KAPR     & \textbf{0.89}  & \textbf{0.98}  & \textbf{0.98}  & \textbf{0.68} & \textbf{0.86}  & \textbf{0.90}  \\ \hline
\multirow{6}{*}{Beauty}      & LVA      & 0.53           & 0.63           & 0.70           & 0.45          & 0.51           & 0.65           \\
                             & Sceptre  & 0.33           & 0.64           & 0.68           & 0.29          & 0.43           & 0.44           \\
                             & SPEM     & \textbf{0.96}  & 0.97           & 0.97           & --             & --              & --             \\
                             & PGPR     & 0.91          & 0.98           & 0.98           & 0.86          & 0.94          & 0.94          \\
                             & DecGCN   &0.89            & 0.97           & 0.97           & 0.87         &  0.95          &  0.95          \\
                             & MFI      & 0.49           & 0.65           & 0.73           & 0.44          & 0.59           & 0.68           \\
                             & KAPR     & 0.94           & \textbf{0.98}  & \textbf{0.98}  & \textbf{0.89} & \textbf{0.95} & \textbf{0.95} \\ \hline
\multirow{6}{*}{Cell Phone} & LVA      & 0.32           & 0.44           & 0.47           & 0.32          & 0.39           & 0.48           \\
                             & Sceptre  & 0.18           & 0.34           & 0.41           & 0.21          & 0.37           & 0.44           \\
                             & SPEM     & 0.56           & 0.60           & 0.62           & --            & --              & --              \\
                             & PGPR     & 0.61           & 0.90           & 0.93           & 0.63          & 0.94          & 0.96           \\
                             & DecGCN   &  0.72          &  0.91          & 0.95           &  0.73        &   0.94         &  0.96          \\
                             & MFI      & 0.22           & 0.39           & 0.49           & 0.39          & 0.57           & 0.69           \\
                             & KAPR     & \textbf{0.78}  & \textbf{0.92}  & \textbf{0.95}  & \textbf{0.75} & \textbf{0.94}  & \textbf{0.97}  \\ \hline
\multirow{6}{*}{Electronics} & LVA      & 0.62           & 0.74           & 0.78           & 0.63          & 0.73           & 0.84           \\
                             & Sceptre  & 0.57           & 0.70           & 0.75           & 0.32          & 0.43           & 0.45           \\
                             & SPEM     & 0.77           & 0.79           & 0.80           & --             & --              & --             \\
                             & PGPR     & 0.81           & 0.91           & 0.94           & 0.50        & 0.88          & 0.93          \\
                             & DecGCN   & 0.79           &  0.82          &  0.87          &  0.45        &  0.82          &   0.91         \\
                             & MFI      & 0.61           & 0.80           & 0.86           & 0.41          & 0.62           & 0.72           \\
                             & KAPR     & \textbf{0.87} & \textbf{0.94} & \textbf{0.96} & \textbf{0.66} & \textbf{0.90}  & \textbf{0.93}  \\ \hline
\end{tabular}
\label{Table 2}
\end{table}
 
From the results in Table\ref{Table 2}, we can draw the following conclusions.

(1) Compared with the embedding-based model (Sceptre, LVA, SPEM, MFI), the graph-based model (PGPR, KAPR) has achieved better results in almost all of the datasets. The main reason is that link constraints between entities ensure they have some share attributes (e.g., same brand, same feature), thus reducing candidate products' search space and improving the accuracy rate.

(2) Compared with the most competitive model PGPR, KAPR has an average relative improvements of 18.0\%, 24.7\% in hits@10 on substitutable and complementary relationship inference. The main reason is that KAPR integrates structured and unstructured knowledge to enhance the discriminability of the model. The dynamic policy network can select actions in a dynamically changing space. This method solves the problem of model training difficulty due to the extensive action space in reinforcement learning.

(3) Compared with the models based only on the product-level feature, the proposed MFI model achieves better effects. MFI has an average relative improvement of 4.11\%, 4.49\% in hits@50 on substitute and complement product inference. The result shows that it is necessary to consider the features of categories, which can solve the data sparsity problem.

\subsubsection{The Ablation Study}
\begin{figure*}[ht]
\centering
\includegraphics[width=0.9\textwidth]{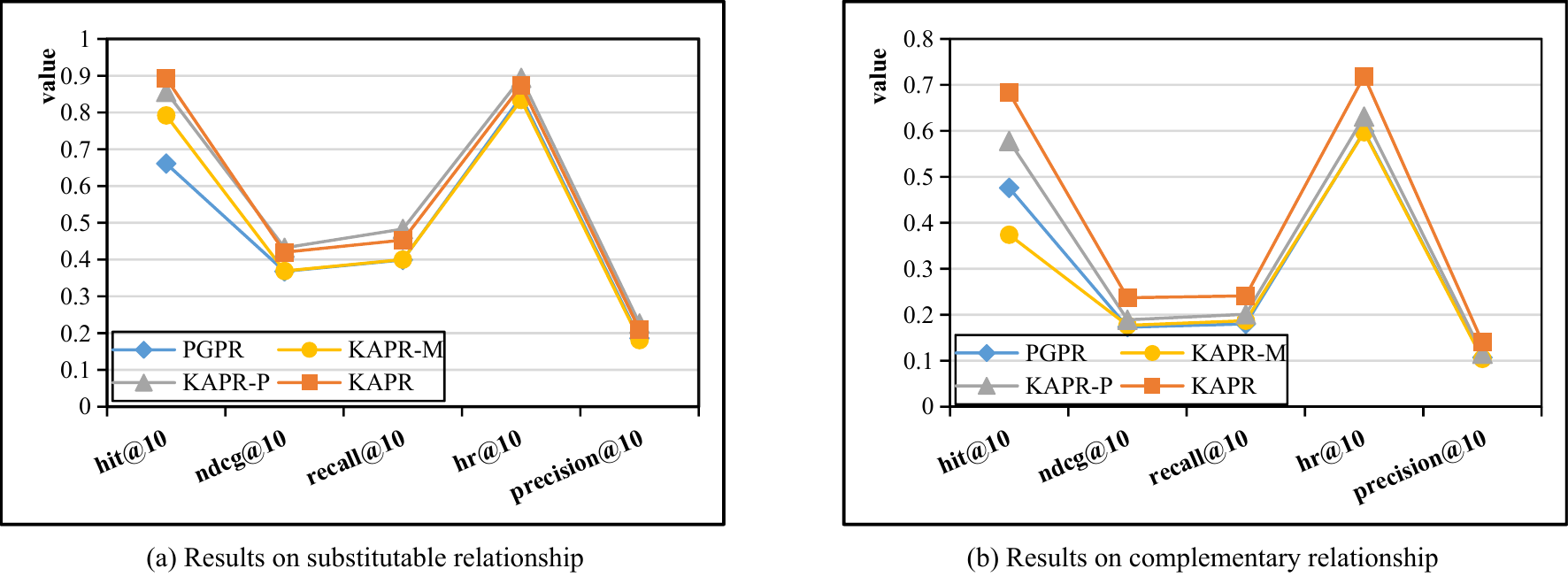}
\caption{Results of ablation experiments on the baby dataset.}
\label{Figure 6}
\end{figure*}

\begin{figure*}[ht]
\centering
\includegraphics[width=0.9\textwidth]{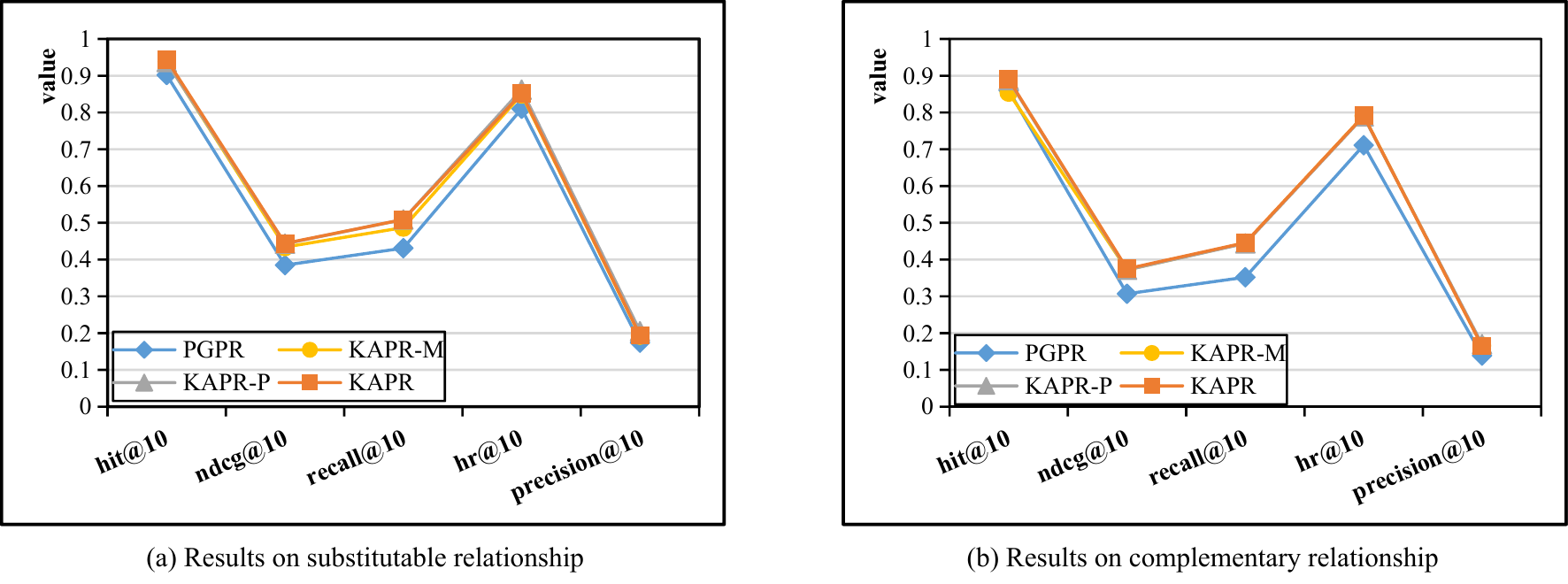}
\caption{Results of ablation experiments on the beauty dataset.}
\label{Figure 7}
\end{figure*}

\begin{figure*}[ht]
\centering
\includegraphics[width=0.9\textwidth]{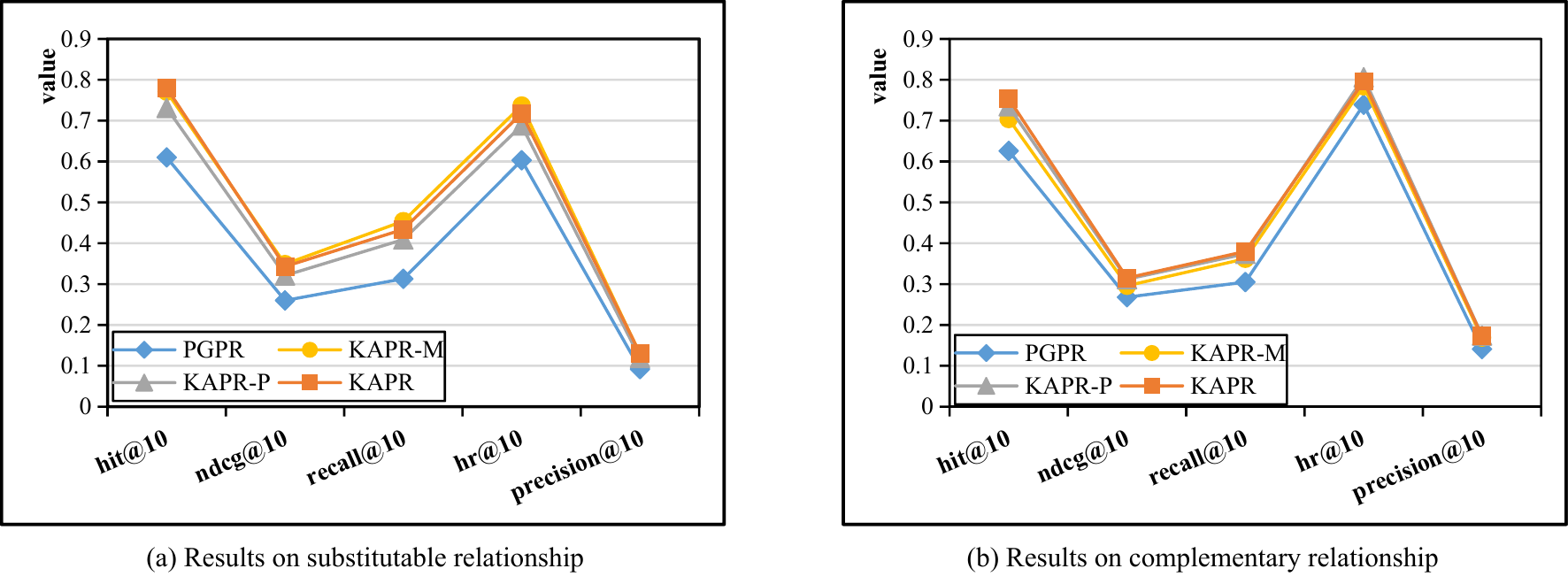}
\caption{Results of ablation experiments on the cell phone dataset.}
\label{Figure 8}
\end{figure*}

\begin{figure*}[ht]
\centering
\includegraphics[width=0.9\textwidth]{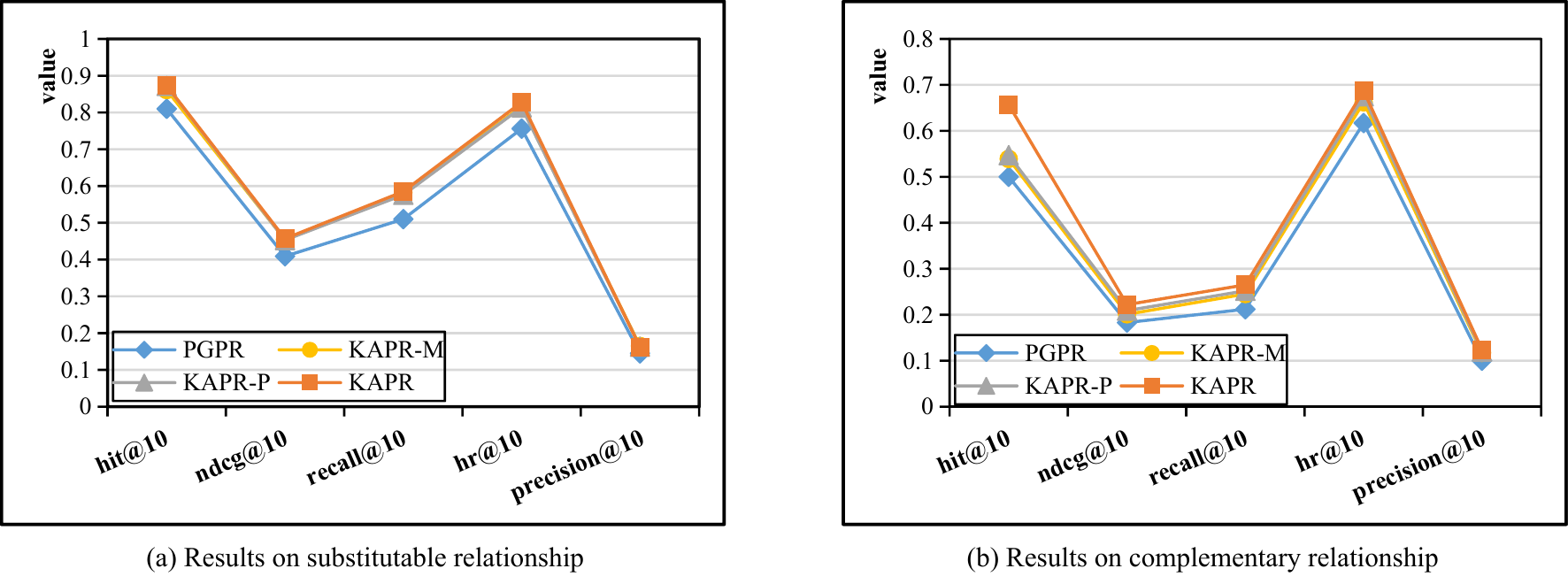}
\caption{Results of ablation experiments on the electronics dataset.}
\label{Figure 9}
\end{figure*}

To further investigate each model component's significance, we conduct ablation experiments on knowledge fusion mechanism and dynamic policy network on four datasets. We are using hit@10, ndcg@10, recall@10, hr@10, and precision@10 as metrics. We design two variants of KAPR as follows. The results are shown in Figure\ref{Figure 6}, Figure\ref{Figure 7}, Figure\ref{Figure 8}, Figure\ref{Figure 9}.
\begin{itemize}
\item KAPR-M: KAPR-M replaces MFI with reward function in the PGPR (TransE).
\item KAPR-P: KAPR-P replaces the dynamic policy network with that mentioned in PGPR.
\end{itemize}
The results shows that the effect of two variant models obtains lower performance on almost all metrics than KAPR. We can observe:
(1)Both KAPR-M and KAPR-P are superior to PGPR for almost all metrics in four datasets. The results show that both the knowledge fusion algorithm and dynamic policy network play an essential role in the inference model. For example, the relative improvements of KAPR-M and KAPR-P over PGPR are at least 22.2\% and 14.1\% on substitutes inference and 6.0\% and 9.0\% on complements inference For all metrics.
(2) Knowledge fusion algorithm and dynamic policy network apply to most datasets. We conduct 8 ablation experiments on 2 relationships of 4 datasets, and each experiment contained 5 evaluation metrics. The investigation finds that the knowledge fusion algorithm and dynamic policy network improve the metrics by 100\% and 87.5\%, respectively, proving that the two components have certain universality.
(3) The reason that KAPR has better effect than KAPR-M and KAPR-P can be summarized as follows.
The result of KAPR-M shows that unstructured knowledge can better judge the relationship than structure knowledge. The reason for the poor effect from KAPR-P is that the dynamic policy network can consider both the state vector and the action vector and selects the best action for the current state through the attention mechanism.

\subsubsection{Visualization of Reasoning}
We analyze the interpretability of the model. We firstly analyze the path pattern output of the model in the reasoning process and then give a case study based on the correct results generated by the model. 

We analyze the validity of the path. We regard paths with length greater than one as valid paths.
\begin{table}
\centering
\small
\caption{Results of the average number of effective paths find in each dataset, the average number of products with a specific relationship find for each product and the average number of paths between a pair of products.}
{
\begin{tabular}{c|c|ccc}
\hline
                                                                                    Category   & Relation & Path/Product & Products & Path/Pair \\ \hline
\multirow{2}{*}{Baby}                                                                  & Sub    & 171.95        & 56.34      & 3.05               \\
                                                                                       & Comp   & 172.37        & 33.18      & 5.19               \\ \hline
\multirow{2}{*}{Beauty}                                                                & Sub    & 169.15        & 69.81      & 2.42               \\
                                                                                       & Comp   & 199.74        & 77.85      & 2.56               \\ \hline
\multirow{2}{*}{Cell Phone} & Sub    & 150.71        & 61.83      & 2.43               \\
                                                                                       & Comp   & 180.81        & 65.11      & 2.77               \\ \hline
\multirow{2}{*}{Electronics}                                                           & Sub    & 131.54        & 51.87      & 2.53               \\
                                                                                       & Comp   & 193.89        & 69.01      & 2.81               \\ \hline

\end{tabular}	
}
\label{Table 7}
\end{table}
As shown in Table\ref{Table 7}, 
we observe that the total sampling paths are 250, and the success rate of effective paths is 0.62 and 0.74. 
The result shows KAPR has good reasoning ability. The average number of related products per product for a relationship is 60. There are about three reasoning paths between each pair of products. The results indicates that the explanations are diverse.
\begin{figure}
\centering
  \includegraphics[width=1\textwidth]{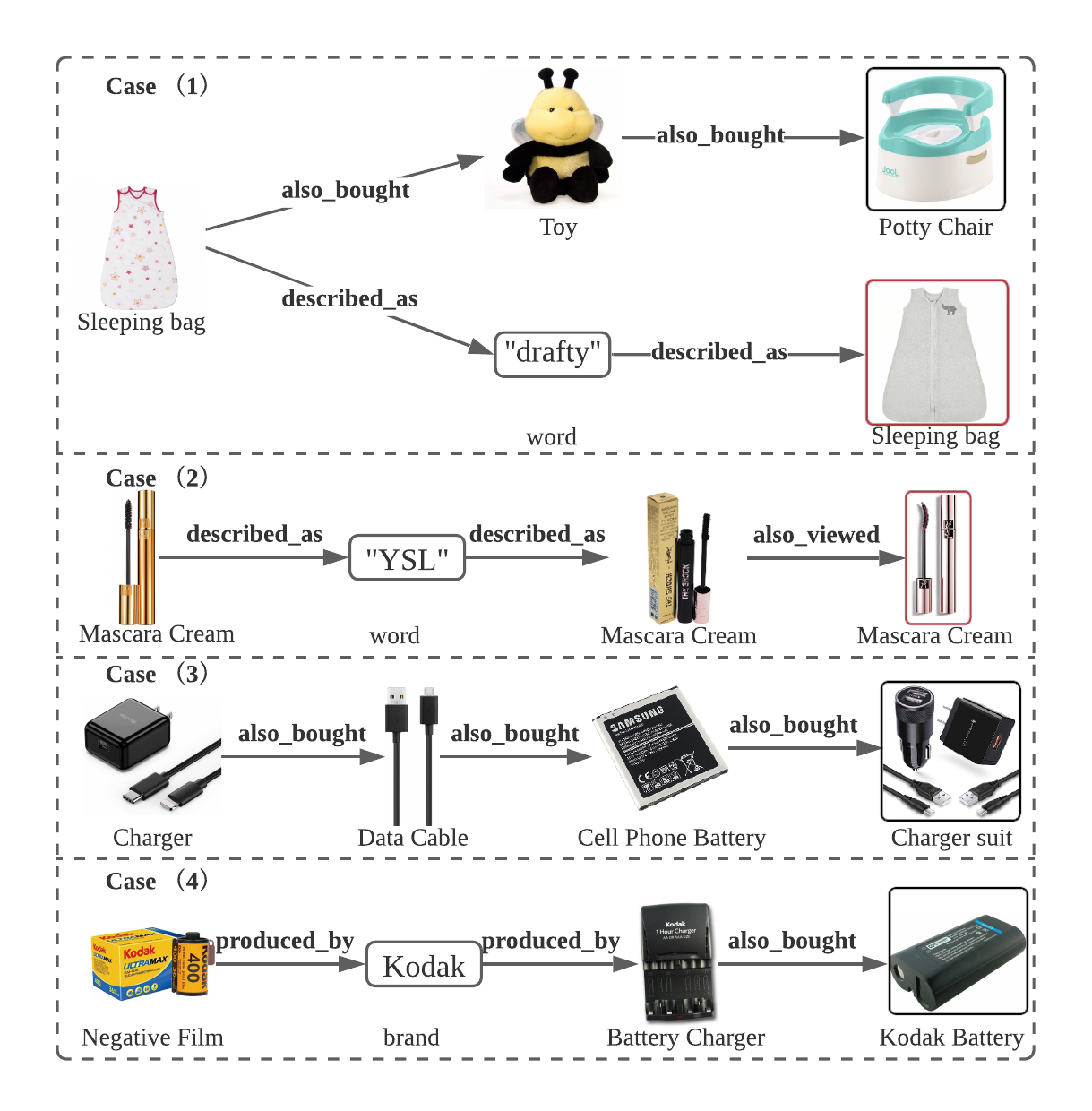}
  \caption{Real cases of reasoning path. The product in the red line represents the substitute, and the product in the black line represents the complement.}
  \label{Figure 4}
\end{figure}
Figure\ref{Figure 4} shows examples of reasoning paths generated by our model. Our model can use rich relationships between entities for reasoning. The relationship between products can be explained by the reasoning path with the attribute node, brand node, or product node. In case 1, we infer that white sleeping bag and pink sleeping bag are complementary products because they had the same attribute (drafty). The potty chair is the white sleeping bag's complement because they have the same complement (toy). In case 2, different mascara cream styles are substitutable products because they belong to the same brand Yves Saint Laurent (YSL). The reasoning path in case 3 links several products (charger, data cable, battery) that are often purchased together to find complementary 'charger suit' of 'charger.' In case 4, the inference path connects several 'Kodak' products as explicit evidence for inference.

\subsubsection{The Effects of Relationships}
\begin{figure}
\centering
\includegraphics[width=1\textwidth]{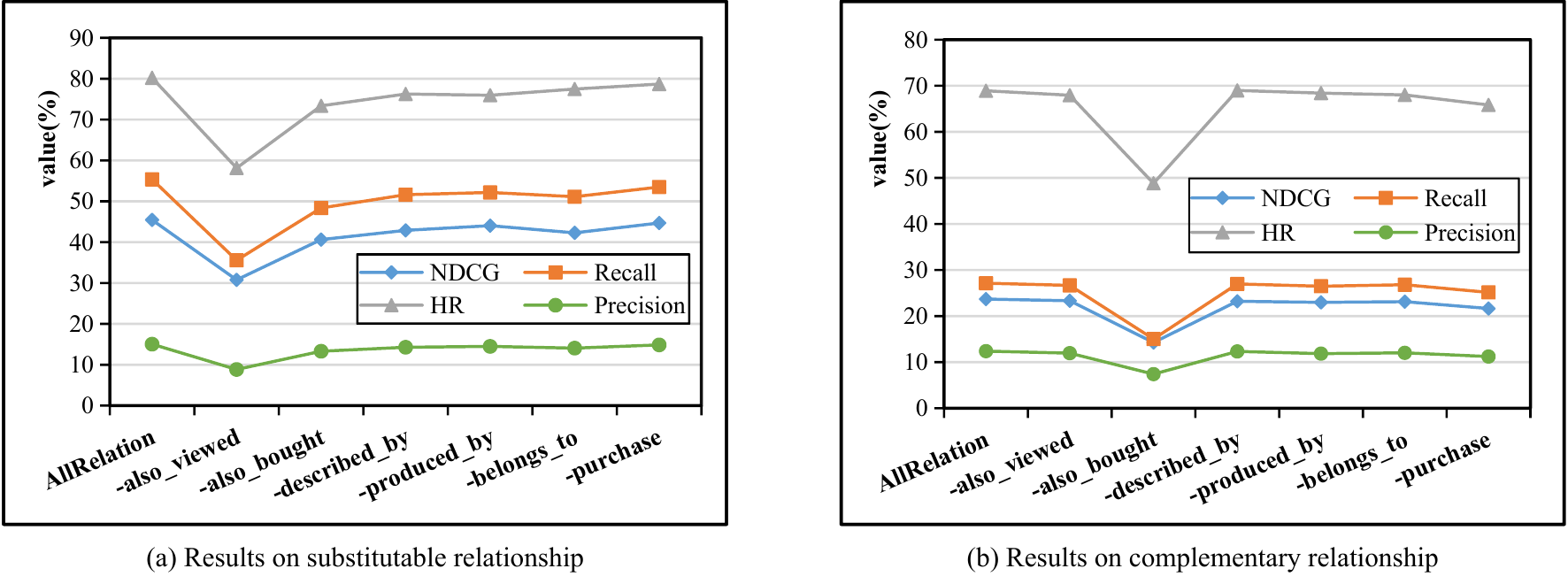}
\caption{The performance of removing each relationship in reasoning stage on electronic dataset for substitutable and complementary relationships.}
\label{Figure F2}
\end{figure}

In this section, we further investigate the significance of each relationship in the reasoning stage. We remove various relationships separately and observe the impact on reasoning.
The results are shown in Figure\ref{Figure F2}, and we have the following conclusions:
(1) In the inference of the two relationships, the model mainly relies on `also\_viewed' and `also\_bought' for reasoning. Other relationships play a smaller role in the experimental results, and removing some edges can improve the experimental results. In future work, the research should focus on optimizing the graph structure.
(2) The lack of a target relationship has the most significant impact on the results. The reason is that the target relationship is the primary edge of the inference meta-paths, and the lack of target relation in the graph leads to the disconnection of part of the meta-paths.
\subsubsection{Sampling Size in Path Reasoning}
We study the influence of sampling size for path reasoning. We design 9 different sampling combinations, and each tuple (N1, N2, N3) represents the number of expansion nodes in each step. The total number of samples for each combination is N1*N2*N3=120 (except for the first case). We test on the Cell phone and Electronics datasets, and the experimental results are as follows. Table\ref{Table Sample} reports the results in terms of NDCG@10, Recall@10, HR@10, and Prec@10.
We observe that the first two levels of sampling sizes play a significant role in reasoning. For example, the result of combination (20,6,1), (20,3,2) is better than combination (10,12,1), (10,6,2). The main reason is that the agent can largely determine the direction of exploration in the first two stpdf, and a larger search space can retrieve more good paths.
\begin{table}[]
\caption{The influence of sampling size at each level in Cell phone on the inference.}
\small
\begin{tabular}{c|cccc|cccc}
\hline
Dataset & \multicolumn{4}{c|}{Substitute}                                   & \multicolumn{4}{c}{Complement}                                   \\ \hline
Sizes   & NDCG        & Recall      & HR          & Prec        & NDCG        & Recall     & HR          & Prec        \\ \hline
25,5,1  & \textbf{0.343} & \textbf{0.434} & \textbf{0.718} & \textbf{0.130} & \textbf{0.315} & \textbf{0.38} & \textbf{0.796} & \textbf{0.173} \\
20,6,1  & 0.303          & 0.366          & 0.643          & 0.108          & 0.221          & 0.258         & 0.670          & 0.115          \\
20,3,2  & 0.262          & 0.296          & 0.562          & 0.085          & 0.223          & 0.249         & 0.658          & 0.110          \\
15,8,1  & 0.322          & 0.401          & 0.679          & 0.117          & 0.295          & 0.347         & 0.759          & 0.147          \\
15,4,2  & 0.277          & 0.317          & 0.607          & 0.091          & 0.292          & 0.327         & 0.729          & 0.133          \\
12,10,1 & 0.320          & 0.400          & 0.671          & 0.118          & 0.287          & 0.339         & 0.745          & 0.145          \\
12,5,2  & 0.274          & 0.321          & 0.592          & 0.093          & 0.274          & 0.309         & 0.707          & 0.123          \\
10,12,1 & 0.330          & 0.413          & 0.699          & 0.122          & 0.287          & 0.339         & 0.742          & 0.147          \\
10,6,2  & 0.273          & 0.323          & 0.594          & 0.093          & 0.256          & 0.301         & 0.696          & 0.120          \\ \hline
\end{tabular}
\label{Table Sample}
\end{table}
\begin{table}[]
\caption{The influence of sampling size at each level in Electronics on the inference.}
\small
\begin{tabular}{c|cccc|cccc}
\hline
Dataset & \multicolumn{4}{c|}{Substitute}                                    & \multicolumn{4}{c}{Complement}                                    \\ \hline
Sizes   & NDCG        & Recall      & HR          & Prec        & NDCG        & Recall      & HR          & Prec        \\ \hline
25,5,1  & \textbf{0.457} & \textbf{0.585} & \textbf{0.828} & \textbf{0.162} & \textbf{0.222} & \textbf{0.265} & \textbf{0.687} & \textbf{0.124} \\
20,6,1  & 0.418          & 0.508          & 0.778          & 0.134          & 0.310          & 0.365          & 0.775          & 0.156          \\
20,3,2  & 0.371          & 0.423          & 0.704          & 0.112          & 0.317          & 0.356          & 0.759          & 0.148          \\
15,8,1  & 0.431          & 0.538          & 0.805          & 0.144          & 0.208          & 0.245          & 0.649          & 0.108          \\
15,4,2  & 0.366          & 0.434          & 0.712          & 0.112          & 0.205          & 0.230          & 0.622          & 0.099          \\
12,10,1 & 0.440          & 0.556          & 0.813          & 0.150          & 0.203          & 0.239          & 0.636          & 0.106          \\
12,5,2  & 0.365          & 0.436          & 0.721          & 0.112          & 0.192          & 0.218          & 0.595          & 0.093          \\
10,12,1 & 0.447          & 0.571          & 0.811          & 0.154          & 0.202          & 0.237          & 0.629          & 0.106          \\
10,6,2  & 0.369          & 0.447          & 0.730          & 0.115          & 0.185          & 0.210          & 0.578          & 0.089          \\ \hline
\end{tabular}
\end{table}
\subsubsection{Influence of Fine-grained Features}
\begin{figure}
\centering
\includegraphics[width=1\textwidth]{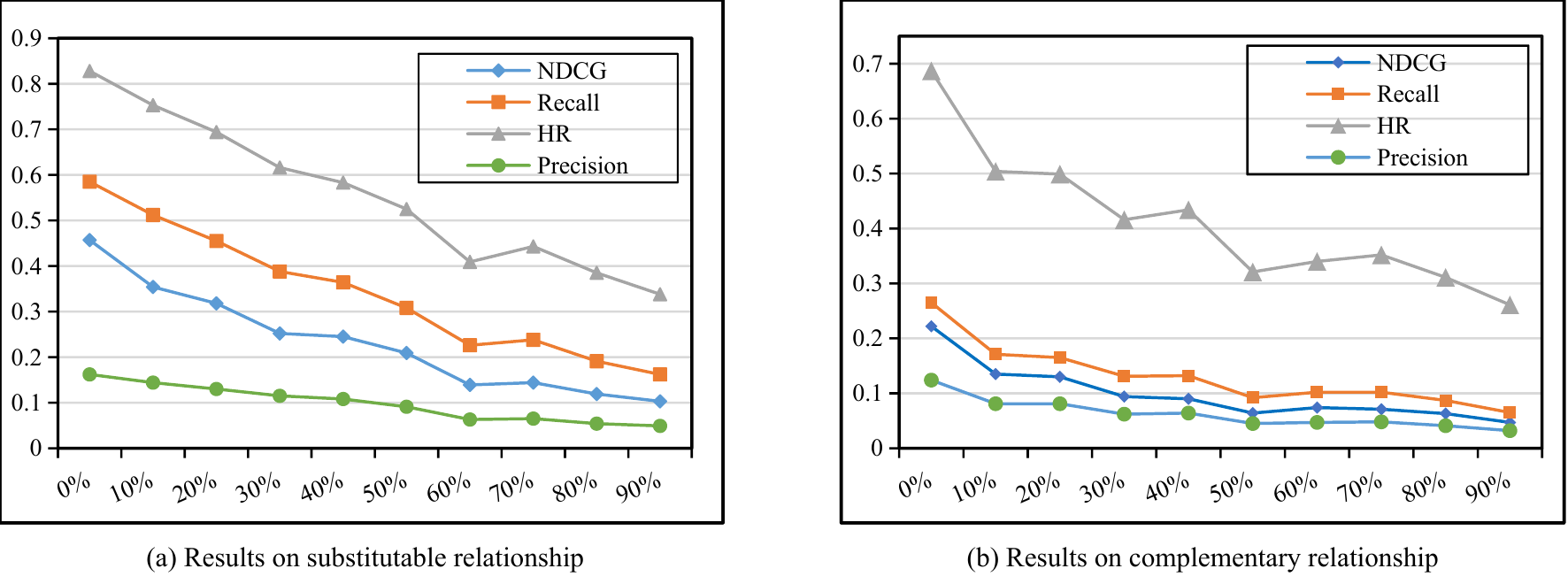}
\caption{The results of removing a specified percentage of the relationship on the electronic dataset.}
\label{Figure F1}
\end{figure}
We design an experiment to verify the influence of the density of fine-grained features on the experimental results. 
Experiments on Electronics dataset, random replace 10\%-90\% relationship, to observe the impact on the experimental results. 
Table\ref{Table F} is the average number of relationships for the head entity after random replacing a certain proportion of the relationships. 
Figure\ref{Figure F1} show the influence of different proportions of fine-grained attributes on the inference of substitutable and complementary relationships. The experimental results show that with the decrease of fine-grained features, the accuracy of the prediction of the two relationships decreases, indicating the important role of fine-grained attributes.
\begin{table}[]
\caption{Number of relations per head.}
\setlength{\tabcolsep}{1mm}
\small
{
\begin{tabular}{l|llllllllll}
\hline
\begin{tabular}[c]{@{}l@{}}random replacement\\ radio\end{tabular} & 0\%   & 10\%  & 20\%  & 30\%  & 40\%  & 50\%  & 60\%  & 70\%  & 80\% & 90\% \\ \hline
described\_by                                                      & 7.17  & 6.11  & 5.39  & 4.73  & 4.05  & 3.36  & 2.74  & 2.01  & 1.36 & 0.70 \\
produced\_by                                                       & 0.69  & 0.62  & 0.55  & 0.48  & 0.41  & 0.34  & 0.28  & 0.21  & 0.14 & 0.07 \\
belong\_to                                                         & 4.39  & 3.94  & 3.50  & 3.07  & 2.65  & 2.18  & 1.74  & 1.29  & 0.87 & 0.44 \\
also\_bought                                                               & 29.68 & 24.93 & 22.17 & 19.41 & 16.63 & 13.84 & 11.11 & 8.33  & 5.56 & 2.78 \\
also\_viewed                                                                & 14.47 & 11.32 & 10.25 & 8.87  & 7.53  & 6.32  & 5.16  & 3.87  & 2.51 & 1.32 \\
purchase                                                           & 44.44 & 40.01 & 35.36 & 31.07 & 26.63 & 22.20 & 17.78 & 13.37 & 8.88 & 4.48 \\ \hline
\end{tabular}
}
\label{Table F}
\end{table}
\subsubsection{Influence of Action Pruning Strategy}
In this experiment, we evaluate the performance of KAPR varies with different sizes of pruned action spaces.
For a given state, we prune actions with the scoring function defined in Equation\ref{eq:score function00}.
The action with a larger score has a greater correlation with the source product and is more likely to be preserved. The purpose of the experiment is to verify whether a larger action space is conducive to finding more accurate reasoning paths. We experiment on Electronics dataset. The pruning space size is set from 100 to 500 with a step size of 100. The results are shown in Table\ref{pruning}. The conclusions are as follows. Our model performance is slightly influenced by the size of the action space. The results firstly increased and then decreased with the size of the pruned action space. The results indicate that when the action space is small, the model can not fully explore. When the action space is too large, many unrelated action nodes are introduced, and the model may learn sub-optimal solutions in a larger space.
\begin{table}[]
\caption{Recommendation effectiveness of our model under different pruned action space sizes on the electronics dataset.}
\setlength{\tabcolsep}{1mm}
\small
\begin{tabular}{l|lllll|lllll}
\hline
Relation          & \multicolumn{5}{c|}{Substitute}                & \multicolumn{5}{c}{Complement}                 \\ \hline
Action space size & 100   & 200            & 300   & 400   & 500   & 100   & 200   & 300            & 400   & 500   \\ \hline
NDCG              & 0.397 & \textbf{0.421} & 0.407 & 0.409 & 0.411 & 0.148 & 0.155 & \textbf{0.162} & 0.153 & 0.161 \\
Recall            & 0.527 & \textbf{0.561} & 0.539 & 0.545 & 0.546 & 0.188 & 0.195 & \textbf{0.199} & 0.195 & 0.201 \\
HR                & 0.774 & \textbf{0.815} & 0.788 & 0.803 & 0.800 & 0.551 & 0.561 & \textbf{0.571} & 0.563 & 0.569 \\
Precision         & 0.147 & \textbf{0.157} & 0.152 & 0.152 & 0.151 & 0.089 & 0.091 & \textbf{0.092} & 0.090 & 0.092 \\ \hline
\end{tabular}
\label{pruning}
\end{table}

\subsubsection{Error analysis}
We observe some error cases of our model and summarize the error reasons, mainly including two points.
(1) Lack of relationships that facilitate reasoning.
In Table\ref{Table 8}, we count the percentages of the correctly and error judged product pairs that have the same node in various relationships.
The percentages of error cases in the `sub' and `comp' are much lower than those of correct issues.
The conclusion is the same as in Figure\ref{Figure F2}, which further illustrates that `sub' and `comp' play an essential role in reasoning.
(2) In some categories, the distinction between the two relationships is not apparent. The concept of substitute and complement is derived from user behaviour.
Products viewed by the same user are called substitutes, and bought together are called complement.
However, some products may have both of the relationships.
For example, two clothes with similar styles may be bought after comparison or purchased together.
Therefore, the boundary between the two relations is relatively fuzzy, bringing specific difficulties to the model's judgment.
\begin{table}[]
\caption{The percentages of the correctly and wrongly judged product pairs that have the same node in various relationships, 33.54 represents 33.54 percent of the correctly judged product pairs connected to the same node through an also\_viewed relationship.}
\centering
\small
\begin{tabular}{c|ccc}
\hline
Relation      & Also\_viewed          & Also\_bought       & Described\_by \\ \hline
Correct\_pair & 33.54        & 20.31      & 50.87         \\
Wrong\_pair   & 3.84         & 1.94       & 17.76          \\ \hline
Relation      & Produced\_by & Belong\_to & Purchase      \\ \hline
Correct\_pair & 1.0          & 5.90       & 10.10         \\
Wrong\_pair   & 1.0          & 1.94       & 0.97          \\ \hline
\end{tabular}
\label{Table 8}
\end{table}
\section{Conclusion}
In this paper, we propose a Knowledge-Aware Path Reasoning (KAPR) to infer the substitutable and complementary relationship, which integrates structured and unstructured knowledge to make the reasoning more robust. 
Based on dynamic policy network with an elegant reward function, our model achieves outstanding performance with explicit inferences. Experiments on four datasets demonstrate the remarkable performance of our model against previous state-of-the-art approaches. Since there exist more fine-grained product relationships among real-world data, we expect this model to be of broad applicability in numerous different tasks of recommender systems. 
\bibliography{aaai21}
\end{sloppypar}
\end{document}